\pdfoutput=1
\documentclass[11pt]{article}
\makeatletter
\def\ps@pprintTitle{%
 \let\@oddhead\@empty
 \let\@evenhead\@empty
 \def\@oddfoot{}%
 \let\@evenfoot\@oddfoot}
\makeatother
\usepackage[utf8]{inputenc}
\usepackage[T1]{fontenc}
\usepackage{fixltx2e}
\usepackage{graphicx}
\usepackage{longtable}
\usepackage{float}
\usepackage{wrapfig}
\usepackage{rotating}
\usepackage[normalem]{ulem}
\usepackage{amsmath}
\usepackage{textcomp}
\usepackage{marvosym}
\usepackage{wasysym}
\usepackage{amssymb}
\usepackage{hyperref}
\usepackage{microtype}
\usepackage[utf8]{inputenc}
\usepackage[T1]{fontenc}
\usepackage{subcaption}
\graphicspath{{figures/}, {all-figures/}}

\usepackage[backref=true,backend=biber,sorting=none,citestyle=numeric-comp]{biblatex}
\hypersetup{colorlinks=true}

\title{High throughput quantitative metallography for complex microstructures using deep learning: A case study in ultrahigh carbon steel}

\author{Brian L. DeCost$^1$ \and Bo Lei$^2$ \and Toby Francis$^2$ \and Elizabeth A. Holm$^2$}

\date{
$^1$Material Measurement Laboratory, National Institute of Standards and Technology, Gaithersburg MD, 20899, USA\\%
$^2$Materials Science and Engineering, Carnegie Mellon University, Pittsburgh PA, 15213, USA\\[2ex]%
\today\\%
{\small Accepted for publication in Microscopy and Microanalysis Aug. 21, 2018 \url{dx.doi.org/10.1017/S1431927618015635}}
}

\begin{document}
\maketitle

\makeatletter
\renewcommand*{\thefootnote}{\@textsuperscript{\tiny{\arabic{footnote}}}}
\makeatother



\begin{abstract}
We apply a deep convolutional neural network segmentation model to enable novel automated microstructure segmentation applications for complex microstructures typically evaluated manually and subjectively.
We explore two microstructure segmentation tasks in an openly-available ultrahigh carbon steel microstructure dataset \cite{uhcsdb,uhcsdata}: segmenting cementite particles in the spheroidized matrix, and segmenting larger fields of view featuring grain boundary carbide, spheroidized particle matrix, particle-free grain boundary denuded zone, and Widmanstätten cementite.
We also demonstrate how to combine these data-driven microstructure segmentation models to obtain empirical cementite particle size and denuded zone width distributions from more complex micrographs containing multiple microconstituents.
The full annotated dataset is available on \url{materialsdata.nist.gov} \cite{uhcsannotations}.
\end{abstract}


\section{Introduction}
\label{sec-1}

Quantitative microstructure analysis is central to materials engineering and design.
Traditionally this entails careful measurements of volume fractions, size distributions, and shape descriptors for familiar microstructural features such as grains and second-phase particles.
These quantities are connected to theoretical and/or empirical models for materials properties, e.g. grain boundary \cite{hall1951} or particle \cite{zener} strengthening mechanisms.
Contemporary microstructure segmentation methods rely on specialized image processing pipelines that often require expert tuning for application to a particular microstructure system.
Furthermore, the microstructures accessible to quantitative analysis are limited by the use of segmentation algorithms that rely on low-level image features (intensity and connectivity constraints).
In this work, we apply deep learning methods for image segmentation to complex microstructure data, with the goal of extending the reach of quantitative analysis to microstructure systems that are currently evaluated subjectively or through laborious manual annotation.

Since 2012, deep learning methods \cite{LeCun2015} have dominated many computer vision applications\footnote{See \cite{Goodfellow-et-al-2016} for a comprehensive introduction to deep learning methods, including architectural and training choices.}, including object recognition and detection, scene summarization, semantic segmentation, and depth map prediction.
The success of deep learning is often attributed to the ability of convolutional neural networks (CNNs) to learn to effectively represent the hierarchical structure of visual data, composing low-level image features (edges, color gradients) into higher level features corresponding to abstract qualities of the image subject (e.g. object parts).
Recently, materials scientists have begun exploring a limited set of applications of contemporary computer vision techniques for flexible and generic microstructure representation.
\cite{decost2015} and \cite{chowdhury2016} explore these techniques in the context of microstructure classification.
\cite{lubbers16_infer_low_dimen_micros_repres} and \cite{decost2017uhcs} use pretrained CNN representations to study relationships between processing conditions and microstructure via dimensionality-reduction and visualization techniques.
\cite{Azimi_2018} use a CNN segmentation model to identify constituent phases in steel microstructures.

In this report, we train a pixelwise CNN \cite{bansal2017} to segment microstructures at a high level of abstraction, and investigate the potential for this technique to enable quantitative microstructure analyses that conventionally would require a large amount of hands-on image processing.
We evaluate the feasibility of this approach on a subset of the openly available Utrahigh Carbon Steel (UHCS) microstructure dataset \cite{uhcsdb,hecht2016,hecht2017}.
CNNs can distinguish between the four principal microconstituents in this heat-treated UHCS: proeutectoid cementite network, fields of spheroidite particles, the ferritic matrix in the particle-free denuded zone near the network, and Widmanstätten laths.
We also train a network to segment individual spheroidite particles, and briefly explore automated microstructure metrology techniques enabled by this kind of powerful segmentation model.
Our training data and annotations for both microstructure segmentation tasks will be publicly available through the NIST materials resource registry \cite{uhcsannotations}.

Our primary contributions are:
\begin{itemize}
\item Establishing two novel microstructure segmentation benchmark datasets
\item Connecting microstructure science to the deep semantic segmentation literature
\item Exploring novel means of expanding contemporary quantitative microstructure measurement techniques to more complex structures
\end{itemize}

For microstructure scientists, CNN-based microstructure segmentation tools require an initial investment in annotation and training, but can enable longer-term or larger-scale research and characterization efforts.
This trade-off is particularly attractive for its potential to enable microstructure-based material qualification by making it easier/cheaper to obtain statistical data on high-level microstructure features known to mediate critical engineering properties of materials (e.g. particle size distributions; denuded zone widths, and particle coarsening kinetics).
In industrial settings where reliance on semi-automated segmentation techniques is common, the barrier to entry is even lower because the training data has already been collected.
CNN-based microstructure segmentation tools also offer a path forward to high-throughput microstructure quantification techniques for accelerated alloy design and processing optimization, where acquisition and analysis of high-quality microstructure data is often a limiting factor.

\section{Methods}
\label{sec-2}
\subsection{Segmentation model}
\label{sec-2-1}

Recently a variety of deep CNN architectures have been developed for dense pixel-level tasks \cite{wang17_under_convol_seman_segmen}, such as semantic segmentation \cite{badrinarayanan2017}, edge detection, depth map, and surface normal prediction \cite{bansal2016marr}.
Conceptually, a modern deep CNN computes a highly nonlinear function through a layerwise composition of convolution, activation, and pooling (i.e. downsampling) functions, the parameters of which are learned from large annotated datasets by some variant of stochastic gradient descent \cite{LeCun2015,Goodfellow-et-al-2016}.
Classification CNNs reduce an input image to a single latent feature vector, where CNNs designed for pixel-level tasks produce a latent representation for every pixel of the input image.
This is typically accomplished by upsampling the intermediate feature maps via a fixed bilinear interpolation \cite{hariharan2015,bansal2017} or a learned deconvolution operation \cite{long2015}.
In the latter class of networks, popular architectures include SegNet \cite{badrinarayanan2017},  Bayesian SegNet \cite{kendall15_bayes_segnet}, U-Net \cite{ronneberger2015} with heavy data augmentation, and fully-convolutional DenseNets \cite{jegou16:_one_hundr_layer_tiram}.
In particular, U-Net \cite{ronneberger2015} was designed for application to medical image segmentation tasks with small dataset sizes, relying on strong data augmentation to achieve good performance.

\subsubsection{PixelNet architecture}
\label{sec-2-1-1}

\begin{figure}
  \frame{
  \includegraphics[width=\textwidth]{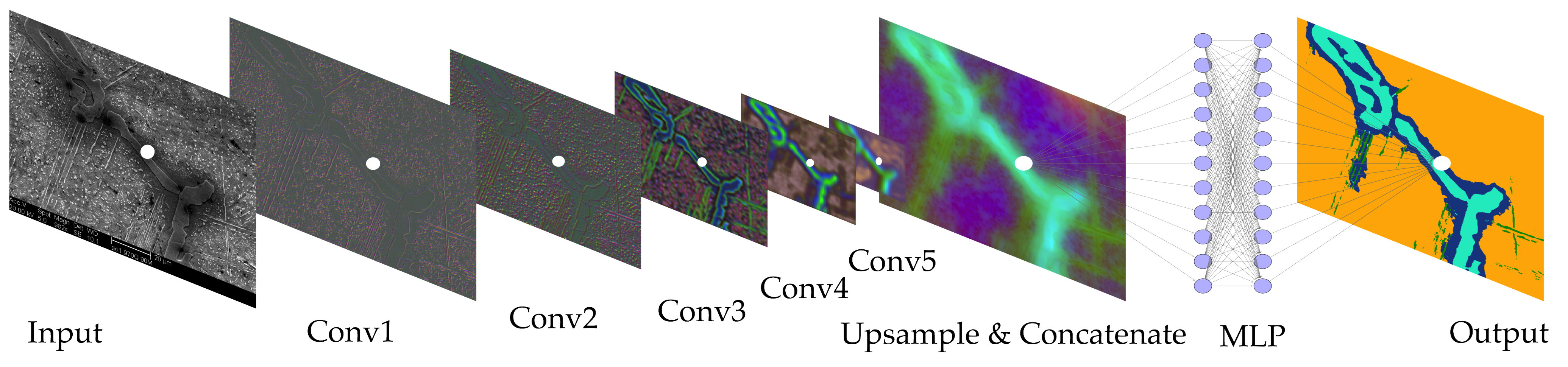}}
  \caption{The PixelNet \cite{bansal2017} image segmentation model. A pixel in the input image (left) is represented by the concatenation of its representations in each convolution layer (white dots). A multilayer perceptron (MLP) classifier is trained to associate the pixel representation with membership in a microstructure constituent (right).}
  \label{fig:architecture}
\end{figure}

The PixelNet \cite{bansal2017} architecture is illustrated schematically in Figure \ref{fig:architecture}.
PixelNet applies bilinear interpolation to intermediate feature maps to form hypercolumn features $h(x) = [conv_1(x),\; conv_2(x),\; \ldots \; conv_5(x)]$, which represent each pixel in the input image with information drawn from multiple scales.
A non-linear classifier implemented as a multi-layer perceptron (MLP, i.e. a traditional artificial neural network (ANN)) maps the hypercolumn features to the corresponding pixel-level target.
Instead of computing dense high-dimensional feature maps at the input resolution as in other popular pixel prediction networks, at training time PixelNet performs a sparse upsampling to efficiently obtain hypercolumn features only for a small sample of the input pixels.\footnote{Our tensorflow implementation of PixelNet is available at \url{https://github.com/bdecost/pixelnet}}
This is attractive for quickly training segmentation networks from scratch with small training sets because it reduces the memory footprint during training and makes training a non-linear predictor with high-dimensional latent representations feasible \cite{bansal2017}.

The feature extraction portion of our PixelNet variant uses the VGG-16 architecture \cite{simonyan2014} used by the original PixelNet \cite{bansal2017}; this architecture consists of 13 convolution layers and two fully-connected layers ${1_1, 1_2, 2_1, 2_2, 3_1, 3_2, 3_3, 4_1, 4_2, 4_3, 5_1, 5_2, 5_3, 6, 7}$.
The MLP layers in our PixelNet variant consist of 1024 neurons with rectified linear (ReLU) activations \cite{nair2010} ($ReLU(y_i) = \max(0, y_i)$ followed by batch normalization \cite{ioffe2015}.
Following the original PixelNet implementation, our hypercolumn features consist of the highest convolution feature map within each block of the VGG architecture (\{1$_{\text{2}}$,2$_{\text{2}}$,3$_{\text{3}}$,4$_{\text{3}}$,5$_{\text{3}}$,7\}), converting layer $7$ to a $7\times7$ convolution filter as in \cite{long2015} and \cite{bansal2017}.
We apply batch normalization \cite{ioffe2015} to each VGG-16 feature map before upsampling via bilinear interpolation, immediately after the ReLU activations.

\subsubsection{Training details}
\label{sec-2-1-2}
We initialize the feature extraction portion of our networks with a pre-trained VGG-16 \cite{simonyan2014} network trained on the ImageNet \cite{Russakovsky_2015} classification dataset.
We train the pixel classification layers from scratch, randomly sampling initial weights from Gaussian distributions with zero mean and standard deviation $\sigma = \sqrt{2/c}$ \cite{he2015}, where $c$ is the dimensionality of the input to the layer.
To prevent overfitting, we use a combination of batch normalization \cite{ioffe2015}, Dropout regularization \cite{srivastava2014}, weight decay regularization \cite{loshchilov17_fixin_weigh_decay_regul_adam} , and data augmentation.
We set the weight decay strength to 0.0005 and apply Dropout regularization with a rate of 10\% after the final MLP layer.
Training images are subjected to local histogram equalization to mitigate differences in overall brightness across different samples and datasets.
The training input and label images are augmented with random rotations in the range $\mathopen[0,2\pi\mathclose)$, horizontal and vertical mirror symmetry, scaling in the range $\mathopen[1,2\mathclose]$, and a \textpm{} 5\% random intensity shift.
Rotated versions of the training input and label images are computed with mirror boundary conditions, with bilinear interpolation for the input images and nearest-neighbor interpolation for the (discrete) label images.
We train the networks with the AdamW optimizer \cite{kingma14_adam,loshchilov17_fixin_weigh_decay_regul_adam} with the recommended default parameters.
First we fix the parameters in the feature extraction portion of the network and train the pixel classification layers with an initial learning rate of 10$^{\text{-3}}$ for 20 epochs (125 gradient updates).
Each gradient update is computed from a random sample of 2048 pixels each from 4 augmented training images.
We then fine-tune the entire CNN for 125 additional gradient updates using AdamW with an initial learning rate of 10$^{\text{-5}}$.

Our dataset has a heavy class imbalance (e.g. Widmanstätten cementite only accounts for $\sim$ 3\% of pixels), so we compare a model trained using the standard cross-entropy classification loss with another trained using the focal loss \cite{lin17_focal_loss_dense_objec_detec}, which is designed for unbalanced datasets.
The focal loss extends the cross-entropy loss function $CrossEntropy(p_t) = - \log(p_t)$, where

\begin{equation}
p_t(p, y) = \begin{cases}
p & \mathrm{if } y = 1 \\
1-p & \mathrm{if } y = 0
\end{cases}
\end{equation}

with ground truth $y$ and predicted class probability $p = P(y = 1)$.
The focal loss adds a modulating factor $(1-p_t)^\gamma$ to emphasize examples about which the classifier is less confident during training, and a scaling parameter $\alpha$ to account for class imbalance:

\begin{equation}
FocalLoss(p_t) = - \alpha_t (1-p_t)^\gamma \log(p_t)
\end{equation}

We follow the recommendation of \cite{lin17_focal_loss_dense_objec_detec} in setting the focusing parameter $\gamma = 2$ and setting the class imbalance parameters $\alpha_t$ proportionally to the inverse frequency of each class.

\subsection{Dataset}
\label{sec-2-2}
The semantic microstructure segmentation dataset consists of 24 manually annotated\footnote{We used the medical image annotation system MITK \cite{mitk}.} micrographs from the open UHCS dataset \cite{uhcsdb,uhcsdata}; examples are shown in Figure \ref{fig:microconstituentresults} and in the online supplemental materials.
These $645 \times 484$ pixel micrographs focus on the characteristic features of heat-treated UHCS: the proeutectoid cementite network and the associated denuded zone, and spheroidized and Widmanstätten cementite.
Multiple heat treatment conditions and magnifications are represented in the semantic microstructure segmentation dataset.

The particle segmentation dataset consists of 24 micrographs collected at a single magnification in support of the particle coarsening analysis reported in \cite{hecht2017}.
Particle annotations were obtained through a partially-automated edge-based segmentation workflow \cite{hecht2017}.
A thresholded blur smooths contrast in the matrix surrounding particles before application of the Canny edge detector \cite{CANNY_1987}.
The particle outlines are filled in, and spurious edges (e.g. at grain boundaries) are removed by a 2px median filter.
The final particle segmentations are verified and retouched manually where the contrast is insufficient for the Canny detector to identify particle edges.
Particles intersecting the edge of the image are removed from the annotations to reduce bias in the estimated particle size distributions.

\subsection{Performance evaluation}
\label{sec-2-3}

Because our set of annotated images is small (24 annotated micrographs total), we use cross-validation to estimate the generalization performance of the PixelNet architecture on our two microstructure segmentation tasks.
We use a 6-fold cross-validation scheme \cite{Hastie_2001}: each dataset is split into six validation sets of four micrographs each, and six PixelNet models are trained on each of the complementary training sets.
The quantitative performance metrics reported in Tables \ref{tab:semanticsegmentationperf} and \ref{tab:particlesegmentationperf} are averages over each validation image in the 6 validation sets; uncertainties are standard errors computed over the six validation images \cite{Hastie_2001}.

We report several standard evaluation metrics for semantic segmentation tasks: pixel accuracy (AC), precision, recall, and region intersection over union (IU) for individual microconstituents.
For each of these metrics, a higher score indicates better performance.

Precision is the fraction of instances predicted to have class $c$ that are correct:

\begin{equation}
Precision(c) = \frac{ \sum_i \hat{y_i} = c \textrm{ and } y_i = c}{\sum_i \hat{y_i} = c}
\end{equation}

where $\hat{y}_i$  indicates the predicted class label for each pixel $i$, and $y_i$  indicates the corresponding ground truth class label.
Equivalently, precision is the ratio of true positives to total (true and false) positives, which decreases when the model overpredicts the number of member pixels in a class.

Recall is the fraction of instances with ground truth class $c$ that are predicted to have class $c$:

\begin{equation}
Recall(c) = \frac{\sum_i \hat{y_i} = c \textrm{ and } y_i = c}{\sum_i y_i = c}
\end{equation}

Equivalently, recall is the ratio of true positives to the total number of pixels in a class, which decreases when the method underpredicts the member pixels in a class.
Since the overall accuracy is defined as the number of true positives divided by the total number of pixels, it is straightforward to show that the classwise average recall or precision equals the overall accuracy.

The intersection over union metric $IU(c)$ for class $c$ (also referred to as the Jaccard metric) is the ratio of correctly predicted pixels of class $c$ (true positives) to the union of pixels with either ground truth or predicted class $c$ (true and false positives plus false negatives):

\begin{equation}
IU(c) = \frac{\sum_i \hat{y_i} = c \textrm{ and } y_i = c}{\sum_i \hat{y_i} = c \textrm{ or } y_i = c }
\end{equation}

For the spheroidite particle segmentation task, we also report performance metrics comparing particle size distributions (PSDs) obtained from the model predictions with those obtained from the ground truth annotations (as reported in \cite{hecht2017}).
We use the two-sample Kolmogorov-Smirnov (KS) test \cite{Massey_1951} to compare each pair of predicted and ground truth PSDs.
The KS score reported in Table \ref{tab:particlesegmentationperf} is the fraction of micrographs where the KS test indicates that the predicted particle size distribution is consistent with the ground truth particle size distribution (i.e. the fraction of micrographs where we fail to reject (at the 95\% confidence level) the null hypothesis that the distributions are equivalent).

\subsection{Computing denuded zone widths \label{sec:dzw}}
\label{sec-2-4}

Given a microconstituent prediction map, we quantify the width of the denuded zone by computing the minimum distance to the network phase for each pixel on the matrix-particle interface.
In practice, we compute a map of Euclidean distance to the network phase, and select the measurements at the denuded zone interface.

To obtain the denuded zone interface, we apply a series of image processing techniques to clean up the microconstituent prediction map, so that only the matrix predictions associated with the diffusion-limited denuded zone adjacent to the proeutectoid cementite network remain.
A morphological filling operation removes any matrix pixels within the network.
Matrix regions that are not connected to the network are identified by application of a morphological closing to matrix phase: any matrix segments that do not intersect the network phase after the morphological operation are removed.
Finally, we remove any matrix predictions that are closer to a widmanstatten region than to a network region, and subsequently remove the widmanstatten regions.
The region boundaries on the cleaned up label image (shown in Figure \ref{fig:denuded_zone}) include only the interface of the proeutectoid cementite network phase (indicated in blue) and the diffuse interface of the denuded zone (indicated in yellow).

\section{Results and Discussion}
\label{sec-3}
\subsection{Semantic microconstituent segmentation}
\label{sec-3-1}

\begin{figure}
  \includegraphics[width=\textwidth]{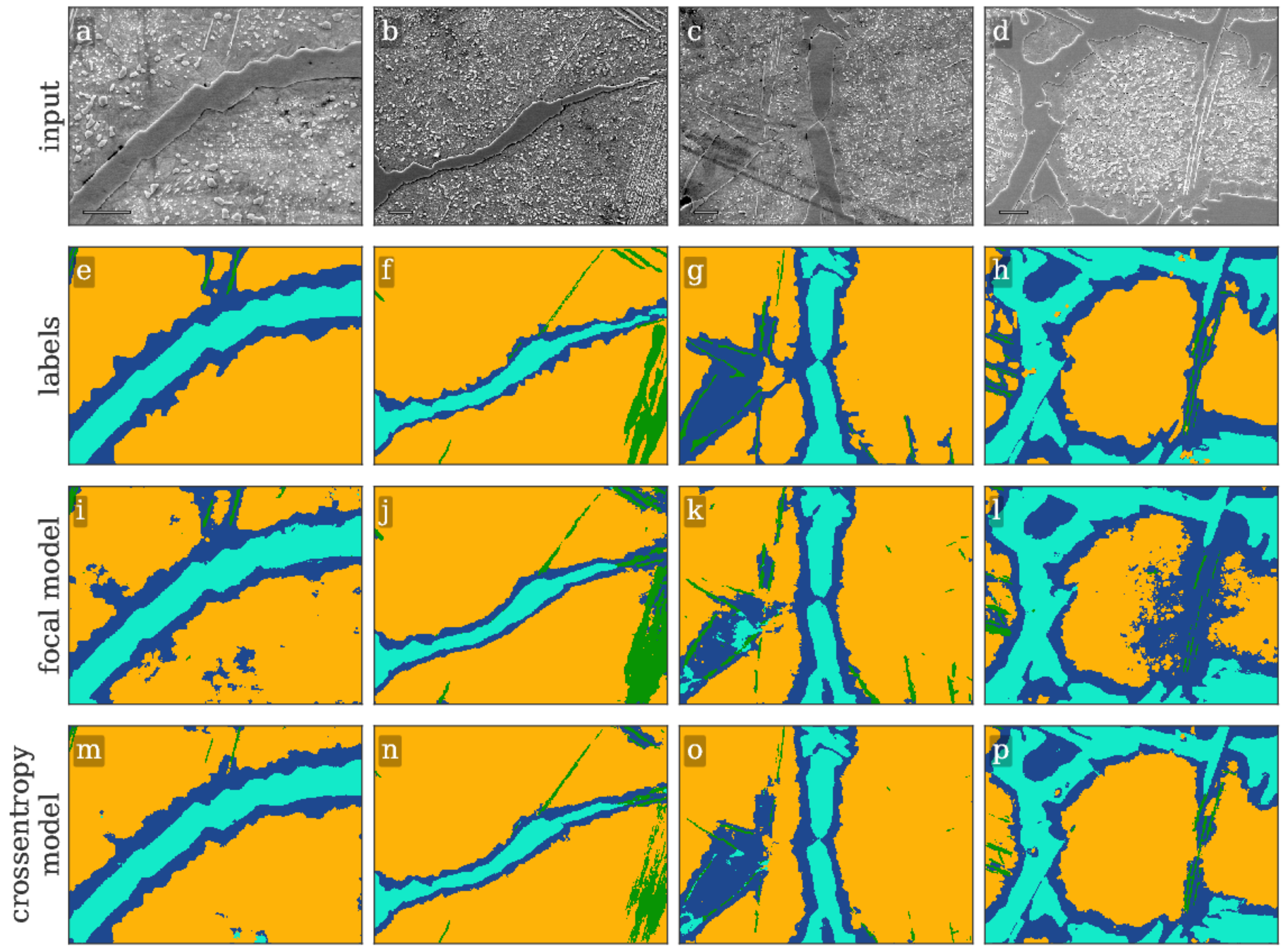}
  \caption{(a-d) Validation set micrographs, (e-h) microconstituent annotations, (i-l) PixelNet predictions using the focal loss segmentation model, and (m-p) PixelNet predictions using the standard crossentropy classification loss. Microstructural constituents include proeutectoid grain boundary cementite (light blue), ferritic matrix (dark blue), spheroidite particles (yellow), and Widmanstätten cementite (green). Scale bars indicate $10 \mu m$.}
  \label{fig:microconstituentresults}
\end{figure}

Figure \ref{fig:microconstituentresults} shows microconstituent annotations (e-h) and predictions (i-p) for the four validation set micrographs (a-d) in one cross-validation iteration; results for all six validation sets are included in the online supplemental materials.
Microconstituent predictions using the focal loss function and the cross-entropy loss function are compared in Figure 2 (i-l) and (m-p) respectively.
The predictions show reasonable correspondence with the annotations, despite nontrivial differences in features, such as particle size and appearance that arise from differences in heat treatment and magnification.
Intensity variations and polishing damage evident in the input images have little impact on the predictive capability of the models.
One notable exception is the cluster of spurious network predictions associated with the damaged areas in the lower left of Figure \ref{fig:microconstituentresults} c.
Both models do a good job respecting the edges of the network carbide phase, with a few exceptions where the network is very fine or the contrast between network carbide and metal matrix is poor (see supplemental Figures S1.1 d and S1.5 d).
Predicted boundaries between spheroidite particles and the denuded zone have little noise and tend to be smoother than in the annotations.
The Widmanstätten predictions show the highest amount of noise, especially where the Widmanstätten lath are fine or are beginning to break up.
The focal loss also tends to surround Widmanstätten cementite with wider swaths of the metallic matrix compared to the annotations.
In addition to the low area fraction of Widmanstätten cementite, one potential contributing factor for these failure modes is labeling bias where the microstructure is ambiguous even to the human expert.
For example, some areas with a low density of spheroidite particles are labeled by the model as metallic matrix where the annotation has made no such distinction.
This phenomenon is evident in the lower half of Figure \ref{fig:microconstituentresults} i, where the model correctly identifies large patches of bare metal in the neighborhood of some large grain boundary cementite particles (refer to supplementary Figure S1.13 a for a more detail).

The cross-entropy model segmentation maps (Figure \ref{fig:microconstituentresults} (m-p)) tend to be more consistent with the annotations for the majority microconstituents.
However, each model errs from the annotations in distinct ways.
In general, the focal loss model seems to emphasize constituent contiguity, while the cross-entropy model tries to resolve fine features.
For example, compared with the annotations in Figure 2 f, the focal loss model (Figure 2 j) more liberally identifies the very fine lath structures in the bottom right corner of the frame, consolidating them into a single patch, while the cross-entropy model (Figure 2 n) produces a noisier map that attempts to track finer-grained details of the lath structure but loses some area fraction.
One conclusion from these results is that while both models give acceptable results, neither is necessarily the best that can be achieved.
For a given segmentation problem, the user must select model parameters and loss functions to achieve the desired quantitative and qualitative performance.

Table \ref{tab:semanticsegmentationperf} shows the average validation set performance with standard errors for the focal loss model.
The model obtains 86.5 $\pm$ 1.6 \% overall accuracy (AC, equivalent to the average of the classwise recall or precision) in reproducing the pixel-level annotations.
The model is consistently good at identifying spheroidite and network regions.
The less prevalent microconstituents (matrix and Widmanstätten) are not as well captured, and show higher variation between images.
For these microconstituents, the recall score is better than the precision score, meaning that the CNN tends to mistake other classes for matrix and Widmanstätten more than it tends to miss genuine matrix and Widmanstätten pixels.
This effect is demonstrated on the fine Widmanstätten lath in the lower right portion of Figure \ref{fig:microconstituentresults} j, where the model fills in the fine spacing between Widmanstätten lath in its prediction.
The low proportion of Widmanstätten pixels in the dataset enhances this effect.
In the case of the matrix class, the difference in recall and precision scores is partly due to the overprediction of metallic matrix in areas containing a low density of spheroidite particles, as discussed in reference to Figure \ref{fig:microconstituentresults} i.

\begin{table}
\caption{\label{tab:semanticsegmentationperf}Focal model semantic segmentation performance averaged over validation images. Uncertainties are standard errors calculated across validation images.}
\centering
\begin{tabular}{lcccc}
 & IU & precision & recall\\
\hline
matrix & 49.1 $\pm$ 3.4 & 60.3 $\pm$ 4.4 & 72.3 $\pm$ 3.7\\
network & 72.9 $\pm$ 5.3 & 85.5 $\pm$ 4.0 & 80.7 $\pm$ 5.9\\
spheroidite & 85.7 $\pm$ 1.8 & 95.1 $\pm$ 1.2 & 89.8 $\pm$ 1.7\\
widmanstatten & 42.7 $\pm$ 2.9 & 50.2 $\pm$ 3.6 & 73.5 $\pm$ 3.9\\
\hline
overall & 62.6 $\pm$ 2.5 & 86.5 $\pm$ 1.6 & 86.5 $\pm$ 1.6\\
\end{tabular}
\end{table}

In contrast, the spheroidite and network classes have slightly higher precision compared with their recall scores.
The standard error for the network scores is large, and is therefore likely accounted for by the small number of gross errors discussed in supplemental Figures S1.1 d and S1.5 d.
Finally, the small difference in precision and recall score for the spheroidite class is likely also due to the overprediction of the metal matrix in regions with low particle density.

Table \ref{tab:crossentropysegmentationperf} shows the same performance metrics for the cross-entropy model trained with revised training hyperparameters.
The crossentropy shows clearly superior overall numerical performance, including a nearly ten point bump in overall IU score.
While most of the per-microconstituent scores are higher for the crossentropy model, the recall score for Widmanstätten cementite is consistently depressed due to underprediction.
We also briefly experimented with the popular U-Net \cite{ronneberger2015} architecture, but found that this architecture slightly underperformed compared with Pixelnet.
These results indicate that various CNN architectures and training schemes can achieve reasonable results, so the user can select an approach based on desired outcomes.
Because of its excellent performance in the spheroidite segmentation task (reported in the next section), we present results only from the Pixelnet model trained with the focal loss function throughout the rest of this manuscript.

\begin{table}
\caption{\label{tab:crossentropysegmentationperf} Crossentropy model semantic segmentation performance averaged over validation images. Uncertainties are standard errors calculated across validation images.}
\centering
\begin{tabular}{lcccc}
 & IU & precision & recall\\
\hline
matrix & 67.3 $\pm$ 6.2 & 81.8 $\pm$ 8.0 & 79.4 $\pm$ 4.0\\
network & 89.4 $\pm$ 3.6 & 96.1 $\pm$ 0.6 & 92.8 $\pm$ 3.5\\
spheroidite & 91.9 $\pm$ 3.5 & 94.9 $\pm$ 2.6 & 96.6 $\pm$ 1.7\\
widmanstatten & 52.9 $\pm$ 6.4 & 72.2 $\pm$ 7.1 & 66.8 $\pm$ 8.1\\
\hline
overall & 75.4 $\pm$ 3.7 & 92.6 $\pm$ 2.5 & 92.6 $\pm$ 2.5\\
\end{tabular}
\end{table}

These quantitative metrics are useful for interpreting the strengths and weaknesses of a particular CNN model, but they do not necessarily directly quantify the quality of the predicted segmentation maps due to inherent subjectivity and bias in the labeling process.
Even a single human annotator will not be able to consistently label an entire dataset, especially for ambiguous higher-level microconstituents such as the  spheroidite class.
For example, the annotator must decide how closely to track cementite particles when tracing out the edge of the denuded zone.
In some cases, it is unclear whether a carbide should be labeled as grain boundary cementite or as a piece of Widmanstätten lath.

Furthermore, the low resolution of the input images relative to some of the finer features of interest also places a practical upper bound on these numerical performance scores, especially for microconstituents with large interfacial areas like the Widmanstätten lath.
Many of the Widmanstätten lath in this dataset are just a few pixels wide, which can lead large shifts in numerical scores for what a human might consider a minor difference in labeling (e.g. dilating or eroding the Widmanstätten lath by one pixel).

\subsection{Spheroidite particle segmentation \label{sec:particles}}
\label{sec-3-2}

\begin{figure}
  \includegraphics[width=\textwidth]{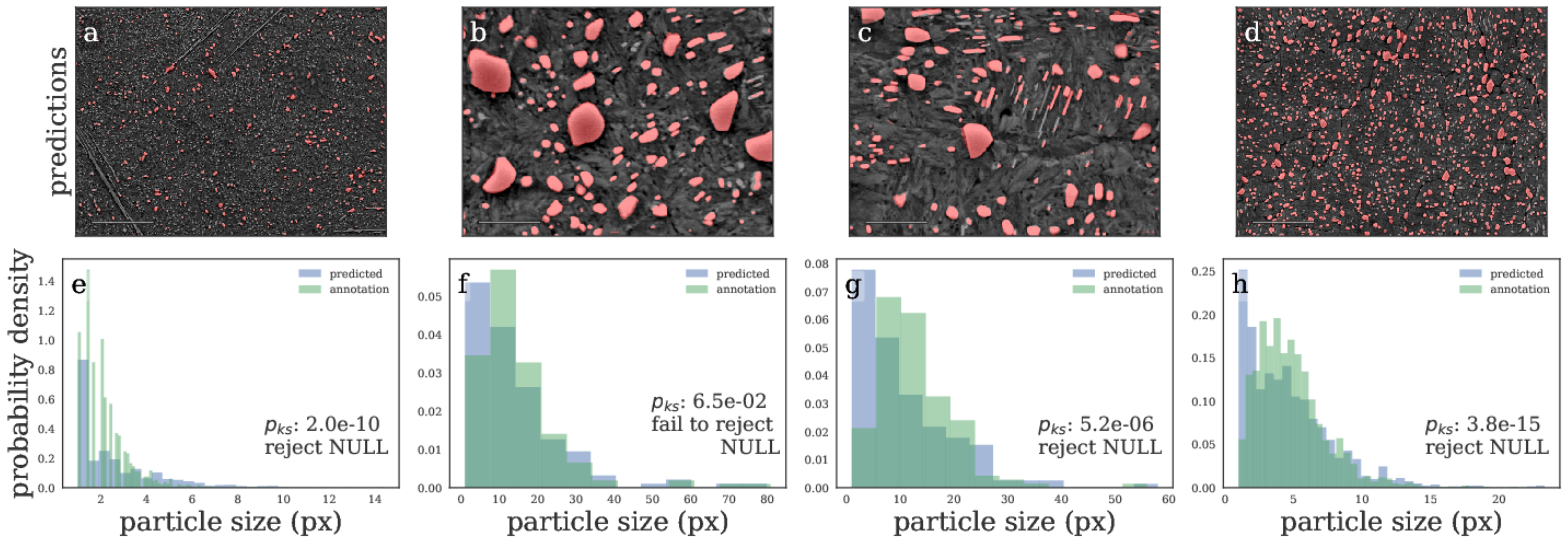}
  \caption{(a-d) Validation set predictions for the spheroidite particle segmentation task, along with (e-h) corresponding derived particle size distributions for the particle predictions (blue) and annotations (green). Scale bars indicate $5 \mu m$.}
  \label{fig:spheroiditeresults}
\end{figure}

Figure \ref{fig:spheroiditeresults} shows some validation results for the individual particle segmentation task, with numerical performance reported in Tables \ref{tab:particleperf} and \ref{tab:particlesegmentationperf}; additional examples are included in the online supplemental materials.
Particle predictions are overlaid in red on the input micrographs (a-d).
The second row (e-h) shows the empirical particle size distributions for both particle predictions and annotations, as well as the results of the two-sample Kolmogorov-Smirnov hypothesis test for distribution equivalence.
Predictions for larger particles relative to the image frame (Figures \ref{fig:spheroiditeresults} b and c) are consistently good, even where contrast gradients across particles and non-trivial background structure challenge thresholding and edge-based segmentation methods.
The primary failure mode of the particle segmentation model is underprediction of very small particles, particularly in Figure \ref{fig:spheroiditeresults} a and d.
The vast majority of the fine particles in Figure \ref{fig:spheroiditeresults} are missing entirely, and many are only partially labeled by the CNN with just one or two foreground pixels.
These particles are typically one to five pixels in size, suggesting that higher- or multi-resolution inputs are necessary for general microstructure segmentation CNNs.
However, the CNN does avoid spuriously labeling the small segments of Widmanstätten in Figure \ref{fig:spheroiditeresults} as particles.

The PixelNet model performs slightly better than Otsu's thresholding method \cite{otsu1979} on all metrics.
One source of bias in these performance measurements are missing particles in the annotations, either from the removal of particles intersecting the image border, or from failure of the semi-automated annotation method itself.
An additional source of bias stems from the application of the watershed algorithm \cite{vincent1991} to split conjoined particles in the annotations; watershed segmentation is not presently applied to the particle predictions, increasing the relative rate of larger particles.

\begin{table}
\caption{\label{tab:particleperf}Particle segmentation performance averaged over validation images. Uncertainties are standard errors calculated across validation images.}
\centering
\begin{tabular}{lcccc}
 & IU & precision & recall\\
\hline
matrix & 90.0 $\pm$ 1.0 & 95.0 $\pm$ 0.6 & 94.5 $\pm$ 1.1\\
spheroidite & 54.8 $\pm$ 3.4 & 74.6 $\pm$ 2.8 & 70.3 $\pm$ 4.3\\
\hline
overall & 72.4 $\pm$ 3.1 & 91.1 $\pm$ 0.9 & 91.1 $\pm$ 0.9\\
\end{tabular}
\end{table}

\begin{table}
\caption{\label{tab:particlesegmentationperf}Particle segmentation performance metrics. Uncertainties are standard errors calculated across validation images.}
\centering
\begin{tabular}{lccccc}
model & IU$_{\text{matrix}}$ & IU$_{\text{spheroidite}}$ & IU$_{\text{avg}}$ & AC & PSD KS\\
\hline
otsu & 86.2 $\pm$ 7.2 & 53.7 $\pm$ 12.1 & 69.9 $\pm$ 9.3 & 88.1 $\pm$ 6.1 & -\\
\hline
pixelnet & 90.0 $\pm$ 1.0 & 54.8 $\pm$ 3.4 & 72.4 $\pm$ 3.1 & 91.1 $\pm$ 0.9 & 0.042\\
\end{tabular}
\end{table}

Despite good numerical performance on the particle segmentation task, the KS test suggests we reject the null hypothesis that the predicted and ground truth particle size distributions are equivalent for all but one of the 24 validation micrographs (shown in \ref{fig:spheroiditeresults} b).
The difficulty in detecting small particles explains the discrepancies between empirical particle size distributions that contribute to the KS score.
For the two validation micrographs in Figure \ref{fig:spheroiditeresults} containing fine particles, the particle size histograms and prediction maps show that the model often entirely misses particles with radii smaller than 5px.
Many of these missing \textasciitilde{}5px particles are partially labeled in the CNN predictions, leading to a severe overrepresentation of single-pixel particles, especially in Figure \ref{fig:spheroiditeresults} h.

\subsection{Quantitative analysis of higher-order features}
\label{sec-3-3}

High-quality automated segmentation techniques for complex microstructure constituents expand the scope of conventional quantitative microstructure analysis by reducing the manual labor required to obtain statistically meaningful amounts of data.
In our UHCS case study, the CNN segmentation model allows us to collect volume and shape statistics for the proeutectoid carbide network, spheroidite particles, and Widmanstätten lath directly from SEM micrographs with no manual intervention.
Additionally, the microconstituent prediction maps enable automated acquisition of interesting microstructural statistics that were previously intractable, such as particle size distributions conditioned on spatial relationships with other microstructure features, or denuded zone widths \cite{hecht2017}.

\begin{figure}
  \includegraphics[width=\textwidth]{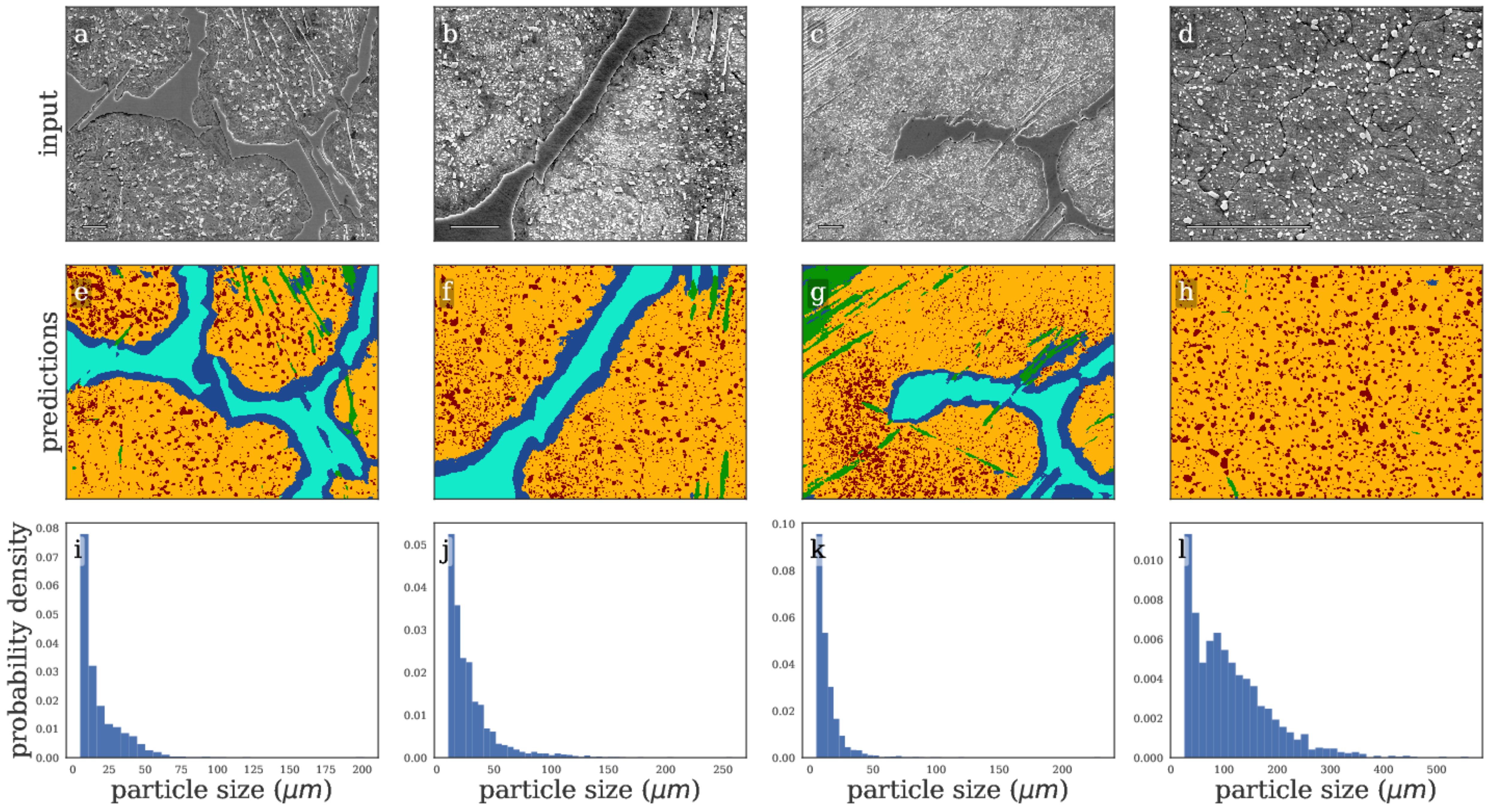}
  \caption{(a-d) Micrographs with (e-h) validation set microconstituent predictions and (i-l) derived particle size distributions obtained by applying the particle segmentation CNN to the semantic microstructure segmentation dataset. Scale bars indicate $10 \mu m$.}
  \label{fig:fused}
\end{figure}

Combining the two microstructure segmentation models allows us to filter out irrelevant microstructure features in order to estimate particle size distributions.
Figure \ref{fig:fused} shows combined microstructure predictions from both the abstract microstructure model and the particle model, using the same color scheme as Figures \ref{fig:microconstituentresults} and \ref{fig:spheroiditeresults}.
We run the input image through separately-trained particle segmentation CNN and microconstituent CNN, suppressing particle predictions (red) outside of the predicted spheroidite regions (yellow).
With an appropriate number of images, one could also compute particle size distributions spatially conditioned on other microstructure features (e.g. distance from the network phase), which could help lead to insights into operative microstructure evolution mechanisms (particle coarsening vs precipitation).
The resolution of these input micrographs is insufficient to yield quantitatively accurate particle size distributions, especially with the underprediction of small particles discussed in Section \ref{sec:particles}, as evident in Figures \ref{fig:fused} b and c.
However, higher quality input and training micrographs will mitigate this effect.

\begin{figure}
  \includegraphics[width=\textwidth]{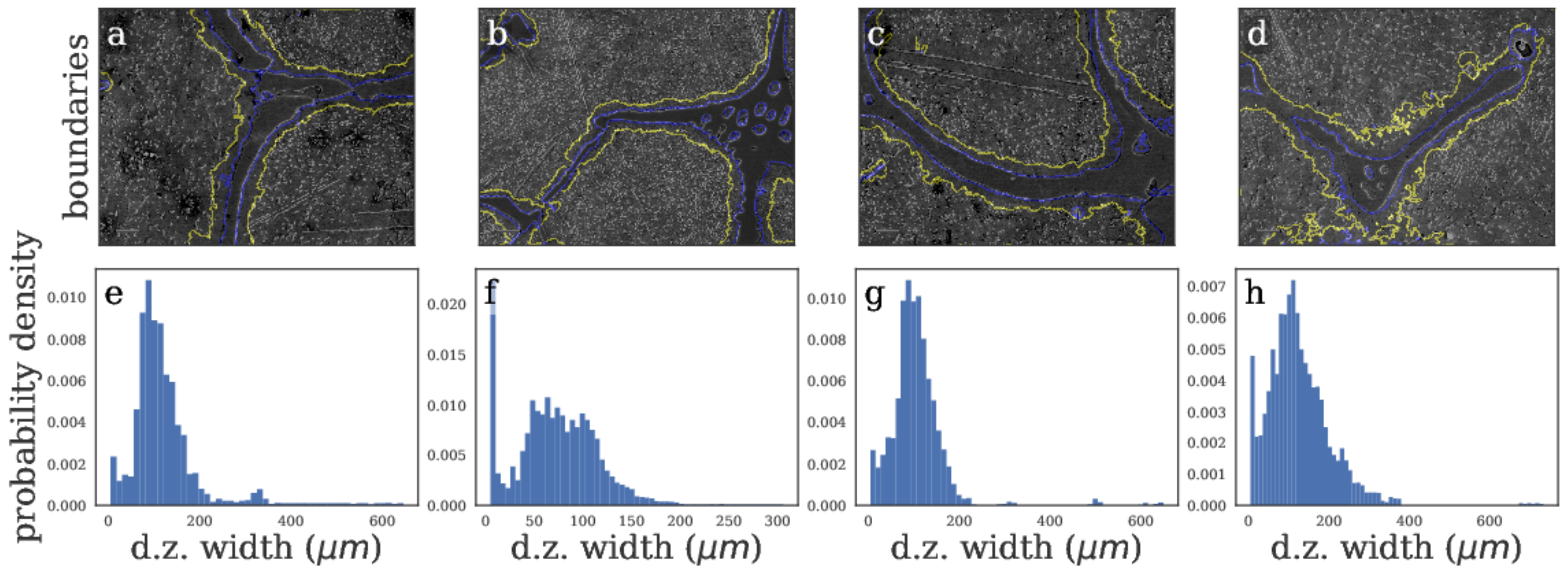}
  \caption{(a-d) Validation set microconstituent predictions with (e-h) corresponding denuded zone width distributions. The network interface is shown in blue and the particle matrix interface is shown in yellow. Scale bars indicate $10 \mu m$}
  \label{fig:denuded_zone}
\end{figure}

Figure \ref{fig:denuded_zone} shows the predicted network and denuded zone boundaries for four validation images with corresponding computed denuded zone width distributions.
The denuded zone width distributions are calculated by aggregating the minimum distance to the network interface for each pixel on the denuded zone boundary, as described in detail in Section \ref{sec:dzw}.
Generally, these empirical denuded zone widths are reasonable, but some care is required to interpret them.
Specifically, the denuded zone width distributions in Figures \ref{fig:denuded_zone} b and d have high frequencies at small spacings that result from spurious cementite network predictions.
Figures \ref{fig:denuded_zone} a and d also exhibit some overprediction of the denuded zone width where the particles are very fine, particularly in the upper portion of Figure \ref{fig:denuded_zone} a.

The initial investment of micrograph annotation and training a CNN makes sense where a statistical number of samples must be characterized in the context of alloy and processing optimization studies, and in the context of microstructure and process validation or verification.
Microconstituent annotation accounted for a substantial portion of the time we spent on this project.
After an initial learning curve, a typical micrograph in our dataset cost between 20 and 30 minutes to annotate.
In contrast, a commodity GPU performs Pixelnet-based segmentation at a rate of approximately one micrograph per second, after an initial training period of two to four hours depending on training hyperparameters.
Success in a practical microstructure science setting will depend on establishing higher-quality training data and deeper understanding of the biases and variance of the labeling process.

The CNN predictions provide some useful feedback on these subjective labeling decisions: consider the micrograph, annotation, and predictions in supplemental Figure S1.6 a, e, and i.
In the bottom half of this micrograph (and in the other micrographs in this validation set), the annotator neglected to label the metal matrix surrounding the Widmanstätten lath as such, while the CNN consistently includes some matrix predictions associated with Widmanstätten predictions.
This subjective labeling decision can be mitigated with higher-fidelity labeling of individual carbide particles -- at much greater labeling expense.
A high quality dataset might be obtained via crowd-sourcing (e.g. students in a microstructure analytics course), generation of realistic synthetic datasets through e.g. phase field modeling, or through the substantial expense of high-resolution elemental mapping with SEM+EDS (Energy-dispersive spectrometry).
A large dataset might also be collected in a semi-supervised fashion through the development of smart microscopes with integrated microstructure recognition features.

Furthermore, it is critical to benchmark microstructure-specific tasks against other popular CNN architectures for semantic segmentation.
Our approach of directly transferring the particle prediction CNN is tenuous, especially due to the disparity in magnification between the general UHCS and specific particle segmentation datasets.
Rather than training two separate CNNs, it may be more appropriate train a single CNN in a multi-task setting, so that microstructures are mapped to a common numerical representation before the respective microconstituent and particle classification tasks.

Finally, microstructure data science is extremely data-limited in comparison to most general computer vision tasks.
Though outside the scope of the present report, a detailed follow-on study to fully characterize the training data requirements of deep learning based microstructure segmentation models would be a valuable tool to enable experimental planning before significant investment for industrial application.
In parallel, collaboration with computer scientists working on low-data deep learning, semi-supervised, and unsupervised techniques could also open the door to applicability in many more microstructure systems, especially where pixel-level annotations are expensive or difficult to consistently obtain.

\section{Conclusions}
\label{sec-4}
We demonstrate microstructural segmentation and quantitative analysis at a high level of abstraction by applying an off-the-shelf deep neural network architecture for pixel-wise prediction tasks.
We also present two new open microstructure segmentation benchmark datasets featuring the microstructures in ultra-high carbon steel at different length scales.
This data-driven approach to microstructure segmentation expands the reach of traditional quantitative microstructure characterization to more complex industrially-relevant microstructure features that have, until now been, difficult to treat in an automated fashion.
Combined with emerging automated microscopy capabilities, data-driven microstructure segmentation systems will enable future applications in high-throughput microstructure studies, including investigations of structure/processing relationships, microstructure design and optimization, and microstructure-based material qualification.

\section*{Acknowledgments}
We gratefully acknowledge funding for this work through National Science Foundation grants DMR-1507830 and CMMI-1826218, and through the John and Claire Bertucci Foundation.
We also recognize the support of the National Institute of Standards and Technology and the National Research Council Research Associate Program.
The UHCS micrographs were graciously provided by Matthew Hecht, Yoosuf Picard, and Bryan Webler (CMU) \cite{uhcsdb}.
Semantic microstructure annotations were performed by B.D.
The spheroidite annotations were graciously provided by Matthew Hecht and Txai Sibley.
The open source software projects Scikit-Learn \cite{sklearn}, scikit-image \cite{walt14_scikit_image}, MITK \cite{mitk}, and keras \cite{keras} were essential to this work.

\printbibliography
\end{document}